\documentclass[
11pt, 
oneside, 
english, 
onehalfspacing, 
headsepline, 
]
{MastersDoctoralThesis} 

\usepackage[utf8]{inputenc} 
\usepackage[T1]{fontenc} 

\usepackage{mathpazo} 

\usepackage[backend=bibtex,style=authoryear,natbib=true]{biblatex} 

\usepackage[autostyle=true]{csquotes} 
\usepackage{geometry}
\usepackage[parfill]{parskip}
\usepackage{amsmath}
\usepackage{tabularx}
\usepackage[ruled,vlined]{algorithm2e}

\thesistitle{Machine learning with limited data} 
\author{Fupin \textsc{Yao}} 
\degree{Doctor of Philosophy}
\subject{Biological Sciences} 
\university{} 
\department{iFlyTek-Surrey Joint Research Centre on Artificial Intelligence, UK} 
\group{{}} 
\faculty{} 

\AtBeginDocument{
\hypersetup{pdftitle=\ttitle} 
\hypersetup{pdfauthor=\authorname} 
\hypersetup{pdfkeywords=\keywordnames} 
}

\begin{document}

\frontmatter 

\pagestyle{plain} 


\begin{titlepage}
\begin{center}

\vspace*{.06\textheight}
{\scshape\LARGE \univname\par}\vspace{1.5cm} 

\HRule \\[0.4cm] 
{\huge \bfseries \ttitle\par}\vspace{0.4cm} 
\HRule \\[1.5cm] 
 
 \large
{\authorname} 

\vfill
\deptname
\vfill
{\large \today}\\[4cm] 
\vfill
\end{center}
\end{titlepage}

\begin{abstract}
\addchaptertocentry{\abstractname} 
Thanks to the availability of powerful computing resources, big data and deep learning algorithms, we have made great progress on computer vision in the last few years. Computer vision systems begin to surpass humans in some tasks, such as object recognition, object detection, face  recognition and pose estimation. Lots of computer vision algorithms have been deployed to real world applications and started to improve our life quality. However, big data and labels are not always available. Sometimes we only have very  limited labeled data, such as medical images which requires experts to label them. In this paper, we study few shot image classification, in which we only have very few labeled data.

Machine learning with little data is a big challenge. To tackle this challenge, we propose two methods and test their effectiveness thoroughly. One method is to augment image features by mixing the style of these images. The second method is applying spatial attention to explore the relations between patches of images.

We also find that domain shift is a critical issue in few shot learning when the training domain and testing domain are different. So we propose a more realistic cross-domain few-shot learning with unlabeled data setting, in which some unlabeled data is available in the target domain. We propose  two methods in this setting. Our first method transfers the style information of the unlabeled target dataset to the samples in the source dataset and trains a model with stylized images and original images. Our second method proposes a unified framework to fully utilize all the data. Both of our methods surpass the baseline method by a large margin. 

\end{abstract}


\tableofcontents 

\listoffigures 

\listoftables 


\mainmatter 

\pagestyle{thesis} 



\chapter{Introduction} 

\label{Chapter1} 

\section{Introduction}
Nowadays, it’s well known that deep learning is powerful. We use it to solve lots of problems in computer vision, natural language processing, and speech processing.  However, we also know that deep learning models have a tremendous number of parameters and consume enormous amount of data. For example, an image recognition model, AlexNet(\cite{NIPS2012_4824}), was trained on ImageNet(\cite{imagenet_cvpr09}), which has millions of images. BERT(\cite{devlin2018bert}), with 340 million parameters, used BooksCorpus (800M words) (Zhu et al.,2015) and English Wikipedia (2,500M words). GPT-3 (\cite{brown2020language}), a recently developed natural language model, has 175 billion parameters. Thus a large number of memory cards, storage drives, and GPUs are consumed. One consequence is that training these large models consumes lots of energy and  produces tons of carbon dioxide, which is not environmentally friendly.

This poses a huge challenge for us. Then can we learn a model with little data? For example, what if you are asked to recognize dogs and cats with only one image labeled as dog and one image labeled as cat? From a statistical point of view, this is almost impossible. Rules can only be derived from a large amount of samples.  But we all know human beings have remarkable ability to learn from little data. For instance, suppose you are given an image of an apple, then you can tell apples of slightly different shape and different color and texture from oranges even though you only see one example of apples.

To address this problem, in recent years, some researchers start to study this few shot learning problem, which is learning with only a very limited amount of data. In this setting, we are allowed to transfer knowledge from another additional dataset and test our trained models on a small dataset. Most few-shot learning methods are also meta learning methods, in which the algorithm tries to learn to learn the model. So it’s called learning to learn, which is a bi-level learning. There are also some non meta-learning based methods, one of which is transfer learning. Using transfer learning, we train a model on the additional dataset and then fine tune it on our target dataset which only has limited data. 

Most of these methods can be categorized as 4 different types of methods: optimization based method, metric learning based methods, data augmentation based methods and parameter generation based methods (or black box / model-based methods). Optimization-based method is the most intuitive meta learning method in which the model tries to optimize the inner optimization problem. The inner optimization tasks are subtasks. By doing this, we hope the model can learn some meta knowledge for all sub-tasks and then generalize to unseen tasks during testing. Metric learning based methods try to find a feature space where all sub-tasks can perform well, i.e., images in the same class are closer in the feature space while images in different classes stay further. Data augmentation based methods deal with the data deficiency by learning to augment the data during meta-training on the meta-training dataset and augmenting the novel dataset during meta-testing. The Parameter generation methods, also called black box or model-based methods, try to generate parameters of sub-tasks from a meta-learner (the learner in the outside level). During training, the model takes samples from a task as input and saves knowledge in the activation states (parameters) in the model. During testing, the model outputs parameters for a task conditioned on the labeled training data and makes predictions using these parameters.

We introduce 2 methods from different perspectives.  Our first style mix method borrows ideas from the literature of image style transfer and then tries to augment the data in the feature space by mixing the style from different images. Doing so, we hope to mitigate the problem of scarce data in few shot learning. Inspired by (\cite{wang2018non}), we also try to use attention mechanisms to address few shot image classification. In the method, we let the query images explicitly attend to important parts of images in the support set. The spatial relations among different parts of image are utilized explicitly, which is not the case in ResNet (\cite{he2015eep}) used by most other methods.

We carefully design and conduct our experiments and thoroughly verify our ideas above. But the improvement is not impressive enough. We then find out the reason why: most of meta learning methods don't work well for few shot classification. We do lots of experiments and gather lots of evidence from recently published papers. A paper [Rapid learning or feature reuse] claims that feature reuse is the reason why most optimization-based methods work, by conducting intensive experiments. We also find that pretraining the feature extractor is vital to achieve high accuracy for most methods. Further meta learning only gives 1 or 2 percent of accuracy improvement. Another factor we found is non-episodic training, such as Baseline++ (\cite{chen2019}) and SimpleShot (\cite{wang2019impleshot}), can achieve almost the same accuracy as episodic training, which is the common practise in few shot learning. Additionally, deeper models can also achieve better accuracies (\cite{chen2019}). And if there is a domain shift between the training set and the testing set, the accuracy will drop significantly (\cite{chen2019}). All these evidences show that the current meta-learning methods don't work well for few shot image classification and can’t help much. Later on, after we found this phenomenon, a series of papers, including (\cite{tian2020ethinking}; \cite{dhillon2019}) were published, verifying this idea. All of these show that meta learning reached its bottleneck and we need some fundamental and theoretical breakthroughs. And it may take several decades to see these breakthroughs.

We then shift research focus to related problems, cross domain few shot learning and cross-domain few shot learning with unlabeled data. We observe that models trained on base dataset perform badly when there is a big domain shift between the base dataset and the novel dataset. Thus it’s important to study cross-domain few shot learning, in which we have one or multiple datasets from different domains as training datasets and we will test our model on another different domain. But this is a too challenging task since during training we are not able to touch the target testing domain. So a new and more realistic setting for few shot learning is proposed and studied. We provide some unlabeled target data for training. In most cases, unlabeled data are easy to obtain. For example, we can have tons of chest X-ray images collected from hospitals when they do body checks every day for patients. But collecting labels (canner or not) is not easy since they require medical expertise, which is usually only owned by doctors. So we propose a new setting, in which there is some unlabeled data in the target domain. 

We introduce two methods we propose to tackle this problem. Our first method transfers domain information to the source labeled dataset by style transfer and then trains a model using the stylized images and the images in the original source dataset. We also propose a pseudo-label method to solve this problem: clustering for all unlabeled data and then co-trained the model using both labled source domain data, clustered target domain data and unclustered target data with a contrastive loss. We achieve big accuracy improvements compared to our baseline for these two methods. We also propose another research idea, which combines domain translation based methods and pseudo-label learning methods. We are implementing this and hopefully we get more improvements.

In the future,we hope to keep exploring cross-domain few shot learning with or without unlabeled data. Also we plan to study compositional and causal models, which are critical abilities when there are only very little data. We hope that by developing compostial and causal models, we can really get rid of the need of big data and build extremely efficient models. Also compositial and causal learning are the ways to general artificial intelligence, in my opinion.

\section{Structure of the Confirmation Report}

In chapter 2, we give a detailed survey about this field. We first give a formal definition of few shot learning. Then we show a taxonomy and an overview of previous methods in literature. Details of important papers in this field are also discussed. In the end, we point out problems and challenges for researchers.

In chapter 3, we introduce two algorithms we propose and experiment results and analysis of these two methods. Since we only have very limited data, we first propose a method to increase features. We then show how it works and experiment results of it. Inspired by (\cite{wang2018non}),we propose a method which fully exploits the spatial relations between the query image and gallery images. We present the experimental results of this method.

In chapter 4, we show the necessities to study cross-domain few shot learning and present a new machine learning setting, cross-domain few shot learning with unlabelled data. We claim this is a more realistic and important setting. Several methods are proposed and show excellent results. We will give some preliminary results and show our future directions.

Finally, we discuss open challenges in this field and our future directions in chapter 6. Cross-domain few shot learning needs to be further explored. Models with compositional and causal learning should be proposed to address the data scarcity problem in few-shot learning.

\section{Novel Work Undertaken}

We propose 2 new methods for the conventional few shot learning and systematically experimental their effectiveness.

We point out the problem in the development of meta learning when no one in the research community noticed this.

We propose a new and more realistic cross domain few shot learning setting, which is cross-domain few shot learning with unlabeled data. We develop two methods in this setting and design and conduct experiments for it. Our methods surpass the baseline methods with a large margin.

\chapter{Literature review} 
\label{Chapter2} 

\section{Definition}
In this chapter we give a detailed introduction to few shot learning. Most of the time, few shot learning means few shot classification. We start from a formal definition. 

\textbf{Conventional supervised machine learning} In the conventional supervised machine learning, we have a dataset $D=\{X_j, Y_j\}_{j=1.. N}$, where $\{X_j, Y_j\}$ is a pair of a data point and it’s label. In the image classification problem, we $X_j$ is an image and $Y_j$ is a category. D can be split into a training set $D_{train}$ and a testing set $D_{test}$. The loss function can be cross-entropy loss. Then we optimize the loss with respect to the parameter theta:
\[\theta ^* = \underset{\theta}{argmin} \;L(D; \theta) \]

\textbf{Few shot learning} In few shot learning, our dataset D is very small (less than 30 samples per class in classification problem). In most cases, we only study the case where there are only less than 10 samples per class. This is an ill-posed problem and almost impossible to achieve high performance without utilizing another dataset based on our current understanding of machine learning. So we are allowed to  use another dataset in the hope that we can learn some meta-knowledge from it and transfer it to the small dataset. The additional dataset is called the auxiliary dataset or the base dataset. 

Most methods in literature address few shot learning problems as meta learning problems. Here, let’s formally define meta learning: We have a dataset D, splitted to a training set (called meta-training set or base dataset) $D_{base}$ and testing set (called meta-testing set or novel dataset) $D_{novel}$. We sample lots of few shot learning tasks from D and assume all tasks have a task distribution $p(\tau)$. During training (also called meta-training), we sample some few shot learning tasks [{$D_{train}$, $D_{test}$] from $D_{base}$. $D_{train}$ and $D_{test}$ are training and testing sets of a few shot learning tasks. They are also called support sets and query sets which are different from meta-training and meta-testing sets. Our learning goal is to minimize a loss function $L(D_{base}, w)$. L measures the performance of the model on a series of tasks. $\omega$ is the parameter of the model, also called meta-parameters which represent meta-knowledge from all sampled tasks:
\[\omega^* = \underset{\omega}{argmin} \;E_{\tau \sim p(\tau)}(D;\omega) \]
During the meta-testing (also called fine-tuning in some papers) stage, we use the meta-knowledge learned during meta-training to learn a model for each task sampled from $D_{novel}$. $\theta$ is the parameter for each few shot learning task:
\[\theta^* = \underset{\theta}{argmax} \;log \;p(\theta|\omega^*, D^{train(i)}_{novel})\]
Finally, we evaluate our algorithm on $D_{novel}^{test(i)}$. Note that $D_{novel}^{train(i)}$ and $D_{novel}^{test(i)}$ are the support and query set of the $i^{th}$ task sampled from the novel dataset.

Meta learning tries to learn meta-knowledge from  a series of few short tasks sampled from the auxiliary dataset and hope the model can perform well to novel tasks which have different classes. In conventional machine learning, we assume there is a data distribution. In meta learning, we assume there is a task distribution. All tasks are sampled from the same task distribution, so if the model fits the tasks sampled from the auxiliary dataset well, it can also generalize to tasks which have new categories. The new categories are never seen by the model during meta-training, so performing well during meta-testing requires generalization.

The above definition is from a task-distribution point of view. There is also another point of view: bi-level optimization point of view (\cite{hospedales2020etalearning}), especially for optimization based methods. This kind of method treats learning to learning as two levels of optimization problem: in the inner loop, you optimize the model for the specific few shot learning problem and get optimal parameters $\theta$ while in the outer loop you optimize the mata-model for all few shot tasks to get the optimal meta-parameters $\omega$. Note that $\theta$ is conditional on the outer-loop parameters $\omega$.  In all optimization based papers, the most famous one is MAML(\cite{finn2017odelagnostic}). In MAML, meta-parameter $\omega$ is the initial value of $\theta$.
\[\omega^* = \underset{\omega}{argmin} \sum_{i=1}^{M} L^{meta}(\theta^{*(i)}(\omega)), \omega, D^{test(i)}_{base})\]
\[s.t. \;\theta^{*(i)}(w) = \underset{\theta}{argmin} \;L^{task} (\theta|\omega, D^{train(i)}_{base})\]
\textbf{Relation between few shot learning and meta-learning} Meta-learning is just one way to deal with few shot learning and it treats few shot learning as a ‘data sample’. There are also other ways to tackle few shot learning, such as transfer learning, such as GPT-3(\cite{brown2020language}).
\section{Relation with other types of machine learning}
\textbf{One/Zero shot learning }When the number of samples in each class is one, it’s called one shot learning. When it’s zero, it’s called zero shot learning. One shot learning is always studied in few shot  learning while zero shot learning is not the same. In few shot learning, we leverage the auxiliary dataset while in zero shot learning, the model relies on attributes or semantic information of those classes to transfer some supervision signals and make learning feasible (\cite{wang2019eneralizing}). There is a chance to borrow some methods from zero-shot learning for Few-shot learning.

\textbf{Transfer learning} Transfer learning tries to transfer knowledge learned from another dataset to the target dataset. In the current setting of few shot classification, it can be viewed as a special case of transfer learning, we transfer knowledge from base dataset to novel dataset.

\textbf{Multitask learning} multitask learning aims to learn several related tasks together. This is a more efficient way of learning so that we can save time and computing resources. In addition, this should also be one feature of general artificial intelligence. Meta learning can also be viewed as a special case of multitask learning in which all tasks are sampled from the same task distribution, which means all tasks share the same statistics. Also during testing, it’s tested on novel tasks. It’s not the case for regular multitask learning.

\textbf{Semi-supervised learning} Semi-supervised learning has both labeled data and unlabeled data while in the classic few shot learning, we only have very limited labeled data. The amount of both labeled and unlabeled data, which semi-supervised data can use, is much more than than few shot learning. There is an intersection for these two types of learning: semi-supervised few shot learning.

\section{Overview of few shot learning}
In this section, we will discuss the taxonomy of few shot learning methods and methods under each category. There are four types of methods: optimization based methods, metric learning based methods, parameter generation based methods (or black box / model-based methods), and data augmentation based methods.

\textbf{Optimization based methods} These methods treat inner-level tasks as optimization problems and try to extract meta-knowledge for inner tasks (\cite{hospedales2020etalearning}). This can also be viewed as learning good global initial points for inner-level tasks so that in the inner loop, after a few steps of gradient descent, the model can quickly adapt to an unseen task. Rapid learning is what we need in mata learning and also avoids overfitting. The most famous one is MAML (\cite{finn2017odelagnostic}), in which in the inner loop, you update the model parameters theta a few steps using $D_{base}^{train}$ and in the outer loop, you update the original model, before being updated by the inner loop, using loss computed on $D_{base}^{test}$. Because MAML differentiates through the optimization process, we need to compute second order derivatives. Then the vanilla MAML algorithm has a great computing cost when there are more than one steps of gradient descent in the inner loop. To mitigate this problem, the author also proposes an approximation method, FOMAML, which drops the second order derivatives and only keeps the first order derivatives. (\cite{rajeswaran2019etalearning}) further proposes a method which only depends on the solution of the inner loop optimization not the path of it, which enables the inner model to update infinite steps without worrying about the computation cost. (\cite{nichol2018n}) presents a method, called Reptile, which keeps sampling tasks and then trains the model on them with a few steps of gradient descent and then moves the meta-parameters towards the updated parameters. LSTM meta learner (\cite{ravi2016optimization}) models the meta-learner as a LSTM. It takes sampled tasks as input and keeps updating the model for T rounds. Their contribution is representing the optimization of the inner task as evolution of LSTM cell states. MetaOptNet (\cite{lee2019etalearning}) uses linear classifiers and SVMs for the base leaner in the inner loop. Since it has a closed form solution, it doesn’t have the high computation cost problem as MAML.

\textbf{Metric learning based methods} try to ‘learn to compare’ (compare samples from query set to the samples from the support set) and obtain a metric function under which data samples in the query and support set stay closer if they are from the same class and stay further if they are from different classes.
\[P_{\theta}(y|x,S) = \sum_{(x_i,y_i)\in S}k_{\theta}(x, x_i)y_i\]
The probability of label y of sample x in the query set is the weighted sum of all the labels in the support set and these weights are generated by a kernel function meaning the distance  between the sample x and  another sample x’ in the support set. (\cite{koch2015siamese}) first use a siamese neural network to deal with this problem. They solve this task as a verification problem in which a pair of images go through the same embedding network and the L1 distance is optimized to be small if they are from the same class. Matching Networks (\cite{vinyals2016atching}), shows the kernel function can be an attention kernel which is the normalized softmax of cosine similarities between samples from the query and the support set. The feature embedding functions for the support and query set are LSTMs. Instead of fixed metric functions, such as L1, L2 or cosine distance , Relation Network (\cite{sung2017earning}) shows that learned metric functions parametrized by neural networks are also possible. Prototypical Networks (\cite{snell2017rototypical}) is a simple metric learning method in which the kernel function is the L2 distance between the query sample and the centroids of all features of  each class. 

\textbf{Model based methods} don’t have an explicit form of the base learner. The meta learner gets samples of each task as input and encodes the information in the internal states of the model. Based on these states, the model makes predictions for the query set.  (\cite{santoro2016meta}) use external memory for encoding and retrieving task information. Because the model is only exposed to the label of $x_t$ at time t+1, the model is forced to predict the label of $x_t$ given data  before time t. (\cite{mishra2017}) uses temporal convolution to gather information for past experience (tasks) and soft attention to pinpoint specific information for the current tasks for prediction. Meta networks (\cite{munkhdalai2017eta}) has meta learner and base learner. The meta-learner is trained to generate parameters (called fast weights in the paper) using LSTM for itself and for the base learner.  (\cite{gidaris2018ynamic}) proposes to leverage an attention base classification weight generator to generate weights for those classes in the base learner. The base learner is a cosine distance classifier. Different from other papers, the author claims that their methods not only perform well on novel classes but also on base classes. Imprinted weights (\cite{qi2017owshot}) has the similar idea that the weight of novel classes are generated using the normalized averaged features of samples in the support set.

\textbf{Data augmentation based methods} mitigate the data scarcity problem by augmenting the data, which is an intuitive strategy. Sometimes, this method can be integrated with other types of methods, such as metric learning based methods. Some of these methods learn a data generator using the base dataset and generate data for novel classes during meta-testing. (\cite{hariharan2016owshot}) aims to augment the data using appearance variance in the base dataset. Style transfer can also be leveraged to augment the data for few shot learning (\cite{antoniou2017ata}). {\cite{wang2018owshot}} integrates data generators with meta learning methods directly in one framework. (\cite{zhang2019ewshot}) uses a pretrained saliency model to segment the foreground and background and then combine these foregrounds and backgrounds to get new images. (\cite{chen2019mage}) has an image deformation sub-network which takes a pair of images as input and one for keeping the content and another for diversifying the deformations. The output images will be used to train the model.

\chapter{Methods for Few Shot Learning} 
\newcommand{\keyword}[1]{\textbf{#1}}
\newcommand{\tabhead}[1]{\textbf{#1}}
\newcommand{\code}[1]{\texttt{#1}}
\newcommand{\file}[1]{\texttt{\bfseries#1}}
\newcommand{\option}[1]{\texttt{\itshape#1}}

\label{Chapter3} 
In this chapter we present two methods we propose for few shot learning from different perspectives. And then we show experiments we conduct and conclusions we draw from these experiments. Our first method, style mix, borrowing ideas from style transfer, augments the data in the feature space. Then we present a method exploring the relationship between tasks during training. We apply knowledge distillation techniques for two tasks sharing the same label space and domain adaptation methods for two tasks having disjoint label space. Inspired by the attention mechanism proposed by Non-local neural networks (\cite{wang2018non}), we explore this idea in few shot learning: ask the images in the query set to attend to pixels in the images in the support set. We will introduce each of them in the following sections.

\section{Style mix for few shot learning}
\subsection{Background}
\textbf{Data augmentation} is widely used in machine learning and deep learning. When you have only very little data, overfitting is almost unavoidable. To increase the dataset without labeling, the easiest way is to augment the dataset. Common augmentation techniques including geometric and color transformation, such as relation, random cropping, random easing, horizontal and vertical flipping,  noise injection, as shown in figure 3.1. These are basic techniques and later researchers also developed more sophisticaled techniques, such as feature space augmentation, adversarial training, and GAN data augmentation. Recently some automatically learned augmentation methods are proposed, such as AutoAugment (\cite{cubuk2018utoaugment}) which search optimal combinations of augmentation techniques as hyper-parameter searching. The principle of data augmentation is to preserve the semantic  identity (labels) while modifying other aspects using some human known prior information. For example, we know rotating images will not alter the label of the objects in the image. Some methods apply too much transformation and the resulting images make less sense to humans.
\begin{figure}[th]
	\centering
	\includegraphics[scale=1.0]{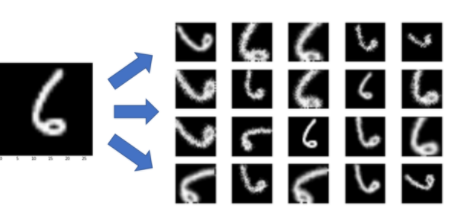}
	\caption{data augmentation (\cite{DataAugm34:online})}
\end{figure}

\textbf{Mixup} is a regulation technique to overcome overfitting in deep learning, first proposed by (\cite{zhang2017ixup}). As shown in the formulas,
\begin{equation}
	\tilde{y} = \lambda y_i + (1-\lambda)y_j, \;where \;y_i,y_j \;are \;raw \;input\; vectors
\end{equation} 
\begin{equation}
\tilde{x} = \lambda x_i + (1-\lambda)x_j, \;where \;x_i,x_j \;are \;raw \;input \;vectors
\end{equation}
 we get the linear combination of a pair of samples and their corresponding labels. Then $\tilde{x}$ and $\tilde{y}$ are used as training samples. This regularization forces models to favor those with linear behaviors of samples. Although it is quite simple, this has been proved to be very effective. Later, in (\cite{verma2018anifold}), instead of using linear combination of samples, it uses linear combinations of hidden representation of samples. They show that this can improve representation by obtaining smoother decision boundaries and fully utilizing hidden representation which contains high level information.
 
 \textbf{Style transfer} is a task in image processing in which the artistic styles of images are required to change but the content should be kept, as shown in figure 3.2. (\cite{gatys2015}) first studies how to use CNN to deal with this problem. In this work, it shows the feature response of a pretrained convolutional neural network can be used as content and the feature summary statistics contains style information. Then by recombining content and style information, we can create new images with different styles. (\cite{dumoulin2016}) shows that a conditional install normalization layer can be used to transfer the styles:
\begin{equation}
	CIN(x,s) = \gamma^s(\frac{x-\mu(x)}{\sigma(x)}) + \beta^s
\end{equation}
 in which the feature x is first normalized by its mean and standard deviation. And then two parameters gamma and beta are learned for the style s. Then (\cite{huang2017rbitrary}) modifies this, and uses the mean and standard deviation of the style image directly without learning those parameters:
 \begin{equation}
	AdaIN(x,y) = \gamma(y)(\frac{x-\mu(x)}{\sigma(x)}) + \mu(y)
\end{equation}
  in which $\sigma$ and $\mu$ are computed across spatial locations of the style image. With this modification, we can replace the style of image x with the style of image y. This method can produce images of  arbitrary styles in real time, so it’s much better compared to the previous method.
  \begin{figure}[th]
  	\centering
  	\includegraphics[scale=0.5]{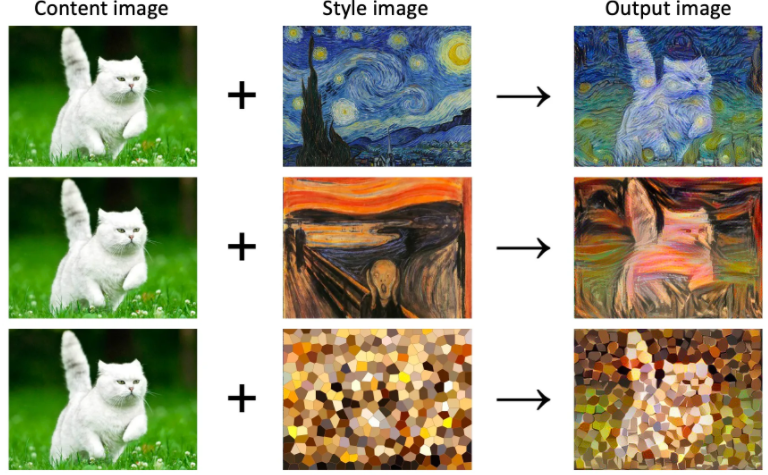}
  	\caption{style transfer (\cite{Howtosty65:online})}
  \end{figure}

\subsection{Approach}
Inspired by Adain (\cite{huang2017rbitrary}) and manifold mixup (\cite{verma2018anifold}), we propose a method to increase the diversity of the features for few shot learning. Because we have very limited data in shot learning, it’s important to augment the data. We propose to mix styles of different images in the same task and in two different tasks. In the first case, during meta learning, for each image, we randomly select another image. Then we compute the mean and standard deviation for these two images across the spatial locations:
\begin{equation}
	\sigma_{nc}(x) = \sqrt{\frac{1}{HW}\sum_{h=1}^{H}\sum_{w=1}^{W}(x_{nchw}-\mu_{nc}(x))^2+\epsilon}
\end{equation}
\begin{equation}
	\mu_{nc}(x) = \frac{1}{HW}\sum_{h=1}^{H}\sum_{w=1}^{W}x_{nchw}
\end{equation}
in which H, W are the height and width of the feature map x. After we get the mean, $\mu$, and standard deviation, $\sigma$ for the two images x and y, we mix the style (styled information is contained in the mean and standard deviation) of them:
\begin{equation} 
	\mu = \lambda*\mu_x + (1-\lambda)*\mu_y 
\end{equation}
\begin{equation} 
	\sigma = \lambda*\sigma_x + (1-\lambda)*\sigma_y
\end{equation}
\begin{equation} 
	\tilde{x} = \sigma(\frac{x-\mu(x)}{\sigma(x)})+\mu
\end{equation}
in which $\lambda$ is the mixing coefficient, sampled from a Beta distribution, which is proposed in (\cite{zhang2017ixup}). Notice in the formula 3.9, we first get the normalized feature without style info and then we scale it with a mixed standard deviation and shift it with the mixed mean. After this operation, the feature will have the mixed style of the images. In this way we hope we get features of diverse styles and we can overcome the data scarcity problem in few-shot learning. The algorithm is shown in Algorithm 1.

\begin{algorithm}[H]
	\KwResult{Classification accuracy}
	\While{Traning}{
		Sample a task T of size n\;
		\While{$n_{image}$<n}{
			Sample two images $x$ and $y$ from T\;
			Get feature maps for $x$ and $y$, f(x) and f(y)\;
			Obtain $\mu_x$ and $\mu_y$ using formula 3.5 and 3.6\;
			Obtain the feature map $\tilde{x}$ with mixed style of image x and y using formula 3.7, 3.8 and 3.9\;
			$n_{image}$ = $n_{image}$ +1
		}
		Calcuate the few shot classfication loss and accuracy using $\tilde{x}$ \;
		Update the model
	}
	\caption{style mix}
\end{algorithm}

\subsection{Experiments}
\textbf{Dataset} We benmark our algorithm on MiniImageNet, which is widely used as a benchmark in few-shot learning. MiniImagenet is a subset of ImageNet (\cite{imagenet_cvpr09}), which has 100 classes in total and 600 images per class. 1000 classes are divided into 64 classes for training, 16 classes for validation and 20 classes for testing. This is a small data but has enough clases and images per class. Thus, it is a perfect testbed for a few-shot learning algorithm. we sample 5-way 1-shot (5 clases and 1 image per class in the query set and 15 images in the support set) tasks and 5-way 5-shot (5 classes and 5 images per class in the query set and 15 images in the support set) tasks for training and testing. During testing ,we average our result over 2000 episodes, while earlier work only use 600 episodes, which has high variance and is not trustable.

\textbf{Backbone} We adopt ResNet12 (\cite{he2015eep}) here for the feature extractor and Prototypical networks (\cite{snell2017rototypical}) for few-shot learning. Earlier work uses Conv4 (Convolutional neural works with only 4 layers) and later on, Resnet12, Resnet18 and wideresnet28-10 are used. Conv4 is too small, has underfit the dataset easily and WideResnet28-10 has too many parameters. We compare Resnet12, Resnet18 and WideResnet and found Resnet12 is enough.

\textbf{Style mix implementation} We implement a stylemix layer in which we perform style mixing within each episode or two episodes. This layer can be inserted after each ResNet block and we tested several combinations.

Training our training has two steps: pretraining on all 64 clases and episodic training using prototypical networks with style mix. PreTraining is widely used in few-shot learning and has been proven to be vital for the final performance. 

\textbf{Result} We give the averaged classification accuracy on the testing set over 2000 episodes in table 3.1.
\begin{table}
\begin{center}
	\begin{tabularx}{0.8\textwidth} { 
			| >{\raggedright\arraybackslash}X
			| >{\raggedleft\arraybackslash}X | }
		\hline
		method   & 5-way 5-shot accuracy \\
		\hline
		Baseline & 78.34  \\
		\hline
		ProtoNet & 79.26  \\
		\hline
		Style mix (after 1st Resnet blcok) & 78.62  \\
		\hline
	\end{tabularx}
\end{center}
\caption{Results of style mix}
\end{table}
In this table, our method achieves comparable results with Protonet. We conduct thorough hyperparameter tuning and detailed analysis trying to figure out why it doesn’t work.

\textbf{Hyperparameter tuning} First we study where we should put the style mix layers and how many style mix layers we should put in the feature extractor. There are 3 possible locations to insert the style mix layer and we also try inserting multiple style mix layers. The result is shown in table 3.2. We can see that the deeper and more the style mix layers are, the worse performance we get.

\begin{table}
\begin{center}
	\begin{tabularx}{0.8\textwidth} { 
			| >{\raggedright\arraybackslash}X
			| >{\raggedleft\arraybackslash}X | }
		\hline
		method   & 5-way 5-shot accuracy \\
		\hline
		Baseline & 78.34  \\
		\hline
		ProtoNet & 79.26  \\
		\hline
		Style mix (after 1st Resnet blcok) & 78.62  \\
		\hline
		Style mix (after 2st Resnet blcok) & 78.47  \\
		\hline
		Style mix (after 3st Resnet blcok) & 77.67  \\
		\hline
		Style mix (after all 4 Resnet blcoks) & 74.35  \\
		\hline
	\end{tabularx}
\end{center}
\caption{Result of style mix in different places}
\end{table}
Then we tune two hyperparameters in our method (table 3.3), $\alpha$ and p. $\alpha$ is a parameter in the Beta distribution, which controls the shape of the distribution. When $\alpha$ gets bigger, we mix the style more. p is the probability of mixing. When p is larger, we mix the styles more. One conclusion can be drawn that no matter how we fine tune these parameters, style mix can not beat the baseline and ProtoNet.

\begin{table}
\begin{center}
	\begin{tabular}{ |p{2cm}|p{2cm}|p{2cm}|p{2cm}|p{2cm}| }
		\hline
		& $\alpha=0.1$ & $\alpha=0.1$ & $\alpha=0.1$ & $\alpha=0.1$ \\
		\hline
		p=0 & & 79.26 & & \\
		\hline
		p=0.2 & & 79.37 & & \\
		\hline
		p=0.5 & 78.87 & 79.26 & 79.37 & 78.60 \\
		\hline
		p=0.8 & & 78.97 & & \\
		\hline
	\end{tabular}
\end{center}
\caption{Hyperparameter tuning for $\alpha$ and p}
\end{table}

\textbf{Analysis} Then we perform some analysis trying to figure out how our method doesn’t work well as we expect. We insert style mix layers only during testing to see the effect and them. As we can see from  table 3.4, the performance gets worse when the style mix layers are inserted deeper. From the below t-SNE embedding of feature maps (figure 3.3), we show that features start to mix together when there is too  much style mixing. So the  discriminative power is reduced. That explain why the stylization doesn’t work.

\begin{table}
\begin{center}
	\begin{tabular}{ |p{2cm}|p{2cm}|p{2cm}|p{2cm}|p{2cm}| }
		\hline
		\multicolumn{5}{|c|}{Insert styleMix during testing ONLY using trained ProtoNet, 5-shot, MiniImageNet} \\
		\hline
		No styleMix & After 1st block & After 2nd block & After 3rd block& After 4th block \\
		\hline
		79.26 & 74.69 & 71.46 & 66.23 & 60.61 \\
		\hline
	\end{tabular}
\end{center}
\caption{Results of style mix when inserting style mix layers during testing}
\end{table}

\begin{figure}[th]
	\includegraphics[scale=0.6]{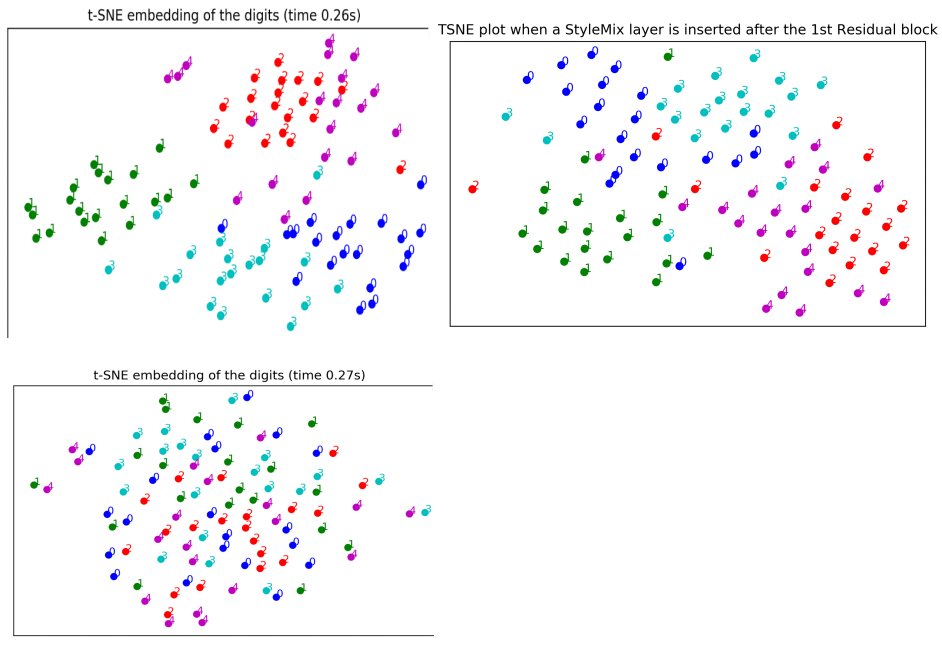}
	\caption{TSNE plots of features}
	Fig 1 (right top): no StyleMix layer. Fig 2 (left bottom): 1 StyleMix after 1st ResNet block. Fig 3 (right bottom): 4 styleMix layers after each ResNet block. Mixing probability is 1.
\end{figure}

\section{Spatial attention for few shot learning}
\subsection{Background}
Attention is all you need (\cite{vaswani2017ttention}) first explores attention mechanism in deep learning. Attention mechanism tries to mimic the attention mechanism in bilgraph, in which we only focus on the most important parts. For instance, if we look at the figure 3.4, most of us only focus or visual attention on the dog head. Then BERT (\cite{devlin2018bert}) and GPT (\cite{brown2020language}) show it’s superior performance in natural language processing. Non-local neural networks is the first work applying attention mechanism in computer vision, in which a spatial attention on feature maps is  designed and shows excellent performance in image and video processing tasks. Then vision transformer (\cite{dosovitskiy2020n}) and i-GPT (\cite{chen2020generative}) shows transformers with attention can even beat Convolutional neural networks when a large amount of data is available.
\begin{figure}[th]
	\centering
	\includegraphics[scale=0.5]{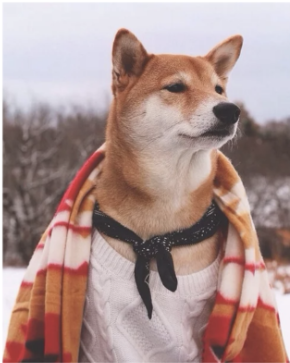}
	\caption{Attention (\cite{Attentio84:online})}
\end{figure}

Here we briefly introduce the most basic self-attention mechanism. It tries to improve the representation of each element in a set by aggregating information from other elements in the same set based on the similarity with other elements:
\begin{equation}
	Attention(Q, K, V) = softmax(\frac{QK^T}{\sqrt{d_k}})V
\end{equation}
The input of a transformer is a set of Q (query), K (key) and V(value). Query, key and value are concepts from retrieval systems. Q is the element we aggregate information for. Key and value are paired. Value is what we look for but it’s often easier to retrieve value by key. In the above formula, we first compute the scaled dot product of Q and K. After applying a softmax function, the attention weight is normalized and it can be viewed as the similarity between the query and the key. based on similarities, we aggregate values. The final output contains information from other elements in the same set so it’s called self attention. And in self-attention, Q=K=V.

Our work is also inspired by DeepEMD (\cite{zhang2020eepemd}) in which image patches are used to computing the matching cost. So the relation between image patches can be utilized.

\subsection{Approach} We develop several variants of our method applying spatial attention in few-shot learning. We start our introduction from the most basic one. We first apply self attention to feature maps of images in the same class. 5 feature maps of the shape NCHW from the same class are concatenated to the shape NC(5H)W. Then we apply self-attention to it. Each element of the feature maps will aggregate information from other elements of the feature maps. Doing so we hope the output feature map becomes more information and then we get better prototypes (averaged feature maps) for late classification. we also apply self-attention to 5 prototypes so that our final feature maps become more separable. The algorithm is shown in Algorithm 2.

Inspired by DeepEMD (\cite{zhang2020eepemd}), we concatenate the feature maps of each query image and 5 prototypes and then apply self-attention. In this way, we hope to fully utilize the spatial correspondence among all locations of the feature map of the query image and 5 prototypes. This is different from the original prototypical networks (\cite{snell2017rototypical}) and here we focus more on local information rather than global information and local information are richer.

\begin{algorithm}[H]
	\KwResult{Classification accuracy}
	\While{Traning}{
		Sample a s-way 5-shot task T of size\;
		\While{$n_{class}$<5}{
			Get five images $x_1$, $x_2$, ... $x_5$ for the class $n_class$ from T\;
			Get feature maps for $x_1$, $x_2$, ... $x_5$, $f(x_1)$, $f(x_2)$, ... $f(x_5)$\;
			concatenate the feature maps $f(x_1)$, $f(x_2)$, ... $f(x_5)$ along the asix H\;
			Applay self attention on the concatenated feature map f(x) using the formula 3.10\;
			Split the feature map after the attention to five feature maps $f(x_1)'$, $f(x_2)'$... $f(x_5)'$\;
			$n_{class}$ = $n_{class}$ +1\;
		}
		Obtain five prototypes by averaging five feature maps in each class\;
		Apply attention on five prototypes\;
		Calcuate the few shot classfication loss and accuracy using augmented prototypes after the attention \;
		Update the model
	}
\caption{Spatial attention}
\end{algorithm}

\subsection{Experiments} We verify our idea on the same MiniImagenet 5-shot taks. Protonet is our baseline. We use ResNet12 as the feature extractor as it’s large enough to fit the dataset compared to Conv4 and less computationally intensive compared to WideResnet-28-10. We implement an attention layer and insert it into different places of the feature extractor, which are layers after each residual block. The result is shown in table 3.5. We can see that no matter where the attention is inserted, the performance doesn’t change. We will discuss the reason in the following section.

\begin{table}
\begin{center}
	\begin{tabularx}{0.8\textwidth} { 
			| >{\raggedright\arraybackslash}X
			| >{\raggedleft\arraybackslash}X | }
		\hline
		method   & 5-way 5-shot accuracy \\
		\hline
		ProtoNet & 79.97  \\
		\hline
		Spatial attention after the 2nd residual block) & 79.59  \\
		\hline
		Spatial attention after the 3rd residual block) & 79.55  \\
		\hline
		Spatial attention after the 4th residual block) & 79.51  \\
		\hline
		Spatial attention after 4 residual blocks) & 80.02  \\
		\hline
	\end{tabularx}
\end{center}
\caption{Results of spatial attention}
\end{table}

\section{Problems of few shot learning methods and remaining challenges}
Since this paper (\cite{raghu2019apid}) was published, we realize that we have reached a bottleneck for all few-shot learning methods. This paper gives a conclusion based on a series of experiments that MAML works is not due to it can rapidly adapt to new tasks and it’s because it learned good features from the base dataset which is useful during testing for the novel dataset. Baseline++ (\cite{chen2019}) shows that a simple cosine classifier can outperform all the star-of-the-art methods. (\cite{tian2020ethinking}) also claims that meta learning doesn’t work well for few shot image classification and all performance boost is from the good feature embedding. (\cite{dhillon2019}) and SimpleShot(\cite{wang2019impleshot}) also show the similar phenomena: feature embedding is the key in few shot image classification. All of this mean that we haven’t got enough progress even though lots of fancy meta learning methods are proposed. The area reaches its bottleneck and we won’t get any real progress until there is a theoretical and funmentai breakthrough. This could explain why our earlier methods don't show significant improvement. It’s hard to get any progress in the area so we transfer our focus to other settings.

\chapter{Cross-domain Few shot learning with unlabeled data} 
\label{Chapter4} 
In this chapter we introduce the domain shift which exists in few-shot learning and two settings of cross-domain few-shot learning. Domain shift is studied in domain adaptation, domain generalization and it has been observed that the accuracy will drop significantly if there exists a domain shift between the base dataset and the novel dataset. To address this problem, we introduce cross-domain few shot learning. This is a more realistic setting as the domain shift is everywhere in our real life. Also, because we know from the previous section that most meta-learning methods just learn useful features, the accuracy will drop if the training and testing distribution doesn’t match. So cross-domain few shot learning becomes an important research problem.

But without touching the target domain where testing happens, the problems become extremely difficult. In most cases, it’s easy to collect some unlabeled data. So cross-domain few shot learning with unlabeled data becomes a more realistic and feasible problem. Studying this problem will have more applications in real life. We will give both the formal definition and related work of cross-domain few shot learning and cross-domain few shot learning with domain shift. 

Then we will introduce two methods we propose for cross-domain few shot learning with unlabeled data. Detailed and through experiment will be shown in later sections.

\section{Background}
\subsection{Domain shift in few shot learning}
Domain shift means the mismatch between two distributions, which is well studied in domain adaptation and domain generalization. Domain adaptation tries to transfer knowledge from one source domain to another domain but these two domains share the same label space. Domain generalization tries to train a model on some source domains and generalize to another unseen domain. The difference between domain adaptation and domain generalization is that in domain generalization the model never sees the target domain during training.

In baseline++ (\cite{chen2019}), the author shows that the classification accuracy will drop if the domain shift between training and testing increases as shown in the figure 4.1. In this paper the author conduct several experiments: evaluating several few-shot learning methods, such as baseline, baseline++, MachingNet, ProtoNet, MAML and RelationNet on CUB->CUB, MiniImageNet, MiniImageNet->CUB. MiniImageNet->CUB means training on MiniImageNet but testing on CUB, which is the cross-domain few shot learning. Domain shift affects the classification accuracy so much and it is so common in real life, so we decide to dive into cross-domain few-shot learning in the next section.

\begin{figure}[th]
	\centering
	\includegraphics[scale=0.6]{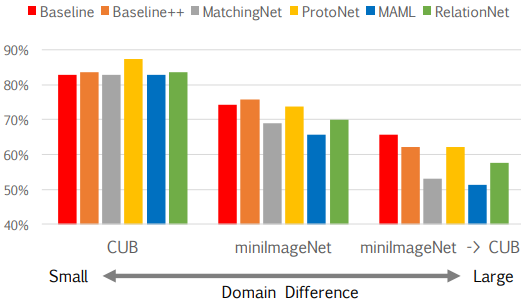}
	\caption{Domain shift in few shot learning (\cite{chen2019})}
\end{figure}

\subsection{Cross domain few shot learning}
\textbf{Definition} We already discuss domain shift in the above section. Here let’s give a formal definition for cross-domain few shot learning:
In cross-domain few shot learning, we have one or several source dataset (base dataset) which has $C1...C_n$ classes and one target dataset (novel dataset) which contains $C_{n+1}...C_{n+m}$ classes. So the source dataset does not contain classes from the novel dataset and vice versa. During training only source dataset can be used and during testing the target dataset is used.

We can see from the above definition that this setting requires generalization abilities which means that the model needs to perform well not only on the base dataset but also on the novel dataset.

\textbf{Related work} There is very little work been done on this topic. Feature-wise transformation (\cite{tseng2020rossdomain}) is one of the first work on this. It tries to augment image features using affine transformation to simulate image distributions from different domains during training. The hyperparameters of the affine transformation are learned using a learning-to-learn approach. In this paper, MiniImageNet, CUB, Cars, Places and Plantae are used as datasets, which are all natural image datasets. This means that the domain shift is not big. Then in (\cite{guo2019}), the author proposes another dataset, which includes non-natural images, such as satellite and chest X-ray images. This makes the benchmark more realistic. The author also proposes some baseline methods in the papers.

But this setting is too challenging as it requires the model to get good performance on unseen domain and unseen classes. So instead we put our focus on another new setting where there are some unlabeled data in the target domain. In this case the model can gather information about the target domain and bridge the domain gap between the base dataset and the target dataset.

\section{Cross domain few shot learning with unlabeled data}
We propose a new setting which is not studied by other researchers before: Cross-domain few-shot learning with unlabeled data. we add unlabeled data for the target domain for two reasons: unlabeled data are easy to collect in most cases and unlabeled data can bridge the domain gap between the source dataset and the target domain.

\subsection{Definition}
We have a labeled dataset from domain S (base dataset) which has classes $C_1...C_n$ and another unlabeled dataset from another domain T containing classes $C_{n+1}...C_{n+m}$ for training. Testing data is from domain T but containing classes $C{n+m}...C{m+p}$. Note that the two later datasets are from the same domain and all 3 datasets have disjoint label space.

\subsection{Related work}
Since this setting is proposed by us, there is no work that has been done by others. Related areas include heterogeneous domain adaptation where the source and target domain don't share the same label space, unsupervised meta learning where the base dataset doesn’t have labels and unsupervised domain adaptation for person/object re-identification.

\textbf{Heterogeneous domain adaptation} is also less studied also. (\cite{gao2020deep}) adopts a deep clustering method for this problem. (\cite{chang2018isjoint}) proposes a method using a shared space between the source and target domain. It trains the model with an unsupervised factorisation loss and a graph-based loss.

\textbf{Unsupervised meta learning} Recently some researchers start to study this problem. UMTRA (\cite{khodadadeh2018nsupervised}) is an unsupervised meta-learning method with tasks constructed by random sampling and augmentation. CACTUs(\cite{hsu2018nsupervised}) uses clustering to automatically construct tasks for unsupervised meta-learning.

\textbf{UDA-Re-ID} is called unsupervised domain adaptation for person/object identification. In this setting, they don’t have class labels but have person ids which can be viewed as fine-grained class labels. The person IDs in the target and source domain are disjoint. There are two types of methods in this field, domain translation based methods ,such as using CycleGan (\cite{zhu2017npaired}) to translate images from the target domain to the source domain and then train a model with translated images. Another type of methods is pseudo-label based method in which clustering techniques are used to extract pseudo-labels for the unlabeled target data and then unlabeled data with pseudo-labels is used to train a model.

\section{Style transfer for cross-domain few-shot learning with unlabeled data}
We propose a method using style transfer. We first feed the images in the unlabeled target dataset as style images and imges in the source dataset as content images to a style transfer model. Then we obtain stylized images which have the same style as the unlabeled target dataset but have same semantic content as the source dataset. We train the stylized images and the original images together. We will show our methods obtain superior results compared to baselines.

\subsection{Background}
Style transfer is introduced in section 3.1.1 so we will not introduce it here. (\cite{geirhos2018magenettrained}) shows that Convolutional neural networks trained on ImageNet are heavily biased toward texture instead of shape, which is not preferred because texture information is not the essential feature of a class. If we change the texture of an object in an image, it should belong to the same class. But the shape is different. Then the authors try to remove texture information and obtain a dataset with only shape information using a style transfer model. Then they use the stylized ImageNet to train a model. Results show that the method is more robust to texture variance and environment changes.

\subsection{Approach}
Our method is inspired by (\cite{geirhos2018magenettrained}). In our case, we have a different problem: cross-domain few-shot learning with unlabeled data. How to utilize the source dataset and the unlabeled target dataset is a question. We come up with a method which transfers the style of the source dataset to the style of the unlabeled target dataset by feeding images in  the unlabeled target dataset as style images and images in  source dataset as content images to a style transfer model (\cite{huang2017rbitrary}). The style transfer model is shown in the figure 4.2. The stylized image looks like images in figure 4.3 if painting images are used as style images and natural images are used as content images.

\begin{figure}[th]
	\centering
	\includegraphics[scale=0.6]{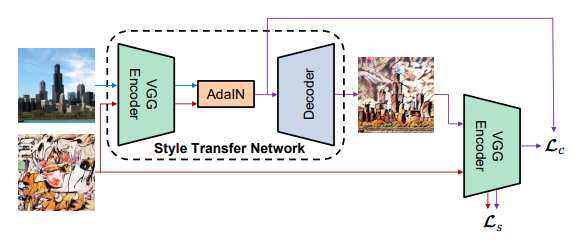}
	\caption{style transfer (\cite{huang2017rbitrary})}
\end{figure}
\begin{figure}[th]
	\centering
	\includegraphics[scale=0.6]{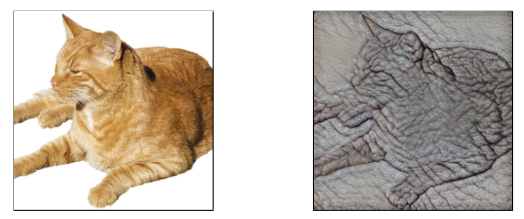}
	\caption{stylized images (\cite{geirhos2018magenettrained})}
\end{figure}

Why is style transfer related to this problem? We find style information carry domain information and lots of style transfer methods share the same ideas as domain adaptation methods. By transferring styles, we actually transfer the domain information and bridge the domain gap between  the source dataset and the target dataset.

After style transfer we train a ProtoNet (\cite{snell2017rototypical}) with both the source dataset images and the stylized images. Testing is conducted on the target testing target dataset. The algorithm is shown in algorithm 3.

\begin{algorithm}[H]
	\KwResult{Classification accuracy}
	Download the style transfer model from (\cite{huang2017rbitrary})\;
	N = total number of images\;
	\While{$n_{img}$ < N}{
		Sample a image $x$ from the source dataset and another image $y$ from the unlabeled dataset\;
		Feed $x$ as the content image and $y$ as the style image to the model and obtain a stylized image $z$\;
		$n_{img}$ = $n_{img}$ +1\;
	}
	Train a ProtoNet with all original images and stylized images
	\caption{Style transfer}
\end{algorithm}

\subsection{Experiments}
\textbf{Dataset} We build our own dataset using DomainNet (\cite{peng2018oment}). The original dataset contains 6 domains: Real, clipart, painting, sketch, infograph and quickdraw and 345 classes. A quick view is in figure 4.4. Instead of using the whole dataset, we select only a subset of the dataset. Since infograph is too noisy and quickdraw contains too little information, we only use four domains: real, sketch, clipart and sketch. Real is used as the source dataset and 3 others are used as target dataset. Similar to MiniImageNet, we have 64 labeled classes as the base dataset and 16 classes for the unlabeled target dataset and 20 classes as the testing target dataset.

\begin{figure}[th]
	\centering
	\includegraphics[scale=0.3]{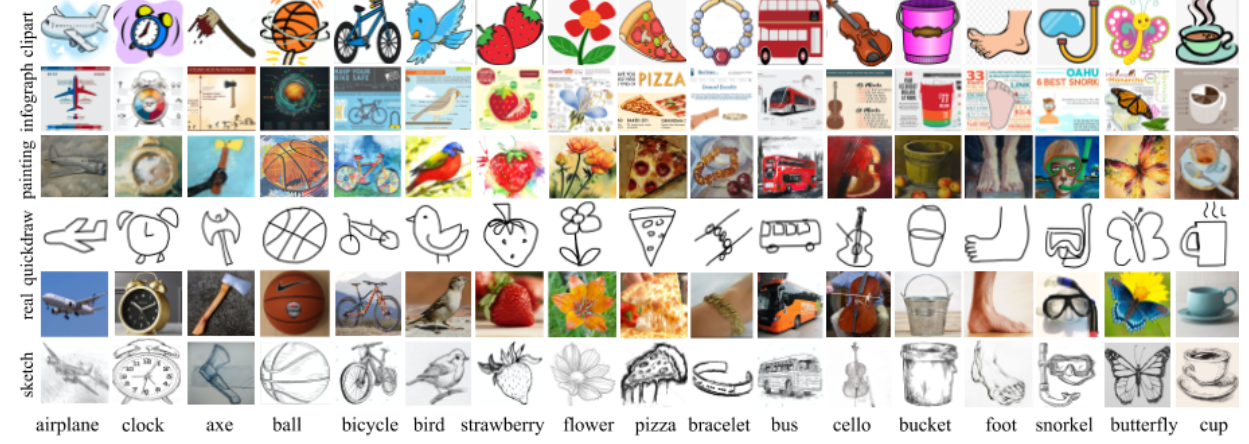}
	\caption{DomainNet (\cite{peng2018oment})}
\end{figure}

\textbf{Result} We still use the same dataset, MiniImagenet to evaluate our algorithm. ResNet12 is used as the feature extractor, the same as before. we also fine-tune the hyperparameter, style transfer coefficient, which means the degree of style transfer. We find we get the best performance when the style transfer coefficient is 1.0. Our result is shown in table 4.1, in which we have two baselines: ‘Backbone trained on real without unlabeled data’ and ‘ADDA’. ‘Backbone trained on real without unlabeled data’ means we only train a ResNet12 using the labeled source dataset without using the unlabeled target dataset and test it on the testing dataset using mean centroid classifier. ‘ADDA’ is a method in (\cite{tzeng2017dversarial}) and we apply it for this problem. The table shows that our stylization method  outperforms all the baseline methods by a large margin, especially on sketch, on which the improvement is 5 percent.

\begin{table}
\begin{center}
	\begin{tabular}{ |p{2cm}|p{2cm}|p{2cm}|p{2cm}|p{2cm}| }
		\hline
		 & clipart & painting & sketch & average \\
		\hline
		ADDA & 59.03 & 55.48 & 53.71& 56.07 \\
		\hline
		Stylization & 67.42 & 60.45 & 58.62 & 62.16 \\
		\hline
	\end{tabular}
\end{center}
\caption{Results of style transfer}
\end{table}

\subsection{Discussion}
Our stylization method has a clear motivation and also shows greater accuracy improvement compared to baseline methods. There is only one defect: the style transfer model used models weights trained on ImageNet, which means that our methods take advantage of another dataset. This is a little bit controversial because we can claim that we only used the style information and the paper, (\cite{geirhos2018magenettrained}), also uses the same method. In the future, we can try to avoid using model weights trained on ImageNet.

\section{Contrastive learning for cross-domain few-shot learning with unlabeled data }
We also propose a method using contrastive learning, in which we use a clustering method, DBSCAN, to get pseudo-labels and update our model using a loss combining all images in the labeled dataset, clustered images in the unlabeled target dataset and unclustered images. We will show that our method obtains surprising results compared to baseline methods.

\subsection{Background}
\textbf{Clustering} is an unsupervised learning method to group samples, which stay close according to some distance measure, together as a cluster when we don’t have labels for it. Common methods are k-means, DBSCAN (\cite{ester1996density}) and Hierarchical clustering. DBSCAN is a density-based method where samples in the high density region are grouped together and points in the low density region are marked as outliers.

\textbf{Contrastive learning} is now widely used in self-supervised learning. In contrastive learning we try to find positive and negative samples and force the model to obtain a small distance between a reference sample and its positive samples and big distance between the reference and negative sample. For example, in SimCLR (\cite{chen2020}), augmented images of the same image are treated as positive samples while other images are treated as negative samples. Using this distance supervision signal we can learn meaningful representation even though we may not have labels. Recent self-supervised methods are going to surpass the performance of supervised methods.

\subsection{Approach}
Our method is inspired by (\cite{ge2020elfpaced}). The original method is designed for object re-identification and we re-purpose it for our setting. First we use the clustering algorithm DBSCAN (k-means can also be applied) to cluster the unlabeled target domain data. Then we have 3 sets of samples: labeled source dataset, clustered unlabeled target dataset and unclustered samples in the unlabeled target dataset. So methods in UDA Re-ID discard those unclustered data but we think it’s a waste of data so we propose a unified framework to fully utilize all the data we have:
\begin{equation} 
	L_f = - log\frac{exp(<f, z^+>/\tau)}{\sum_{k=1}^{n^s}exp(<f, w_k>/\tau) + \sum_{k=1}^{n_c^t}exp(<f, c_k>/\tau) + \sum_{k=1}^{n_o^t}exp(<f, v_k>/\tau)}
\end{equation}

This is our loss function in which f is a feature vector of an image either from the source dataset, clustered unlabeled target dataset or uncluster outliers. $z^+$ indicates a positive prototype corresponding to $f$. $w_k$ is the centroid of all features in the class $k$, $c_k$ is the centroid of all features in the cluster $k$ and $v_k$ is the $k^{th}$ outlier feature. If f is from the source dataset, then $z^+=w_k$ is the centroid feature vector of class k that f belongs to; if f is from the clustered dataset, then $z^+=c_k$ is the centroid feature vector of the cluster $k$ that f belongs to and if $f$ is from the unclusted outliers then $z^+=v_k$ is the outlier feature that f belongs to.

Our algorithm alternates between updating the model using the above comparative learning loss and producing pseudo-labels using clustering methods. Detailed description is show in algorithm 4.

\begin{algorithm}[H]
	\KwResult{Classification accuracy}
	Let N be the number of iterations for updating the model after each clustering\;
	\While{Traing}{
		Obtain pseudo-labels for clustered samples and unclustered samples using DBSCAN algorithm with the unlabeled dataset\;
		$iter$ = 0;
		\While{$iter$<N}{
			$iter$ = $iter$ + 1 \;
			Updating the model using the loss function in formula 1.1 with all labeled samples in the source dataset, clustered samples with their pseudo-labels and unclustered samples\;
		}
	}
	\caption{Contrastive learning}
\end{algorithm}

\subsection{Experiment}
We first benchmark our algorithm on the same dataset as the dataset in 4.3. Resnet12 is used as the feature extractor. we update the model 400 iterations after each step of clustering. As clustering takes a long time (several minutes), we should update our model for more iteration after each clustering. Our result is shown in table 4.2:

\begin{table}[h!]
\begin{center}
	\begin{tabular}{ |p{5cm}|p{1.5cm}|p{1.5cm}|p{1.5cm}|p{1.5cm}| }
		\hline
		& clipart & painting & sketch & average \\
		\hline
		Backbone trained on real without unlabeled data & 65.17 & 59.13 & 53.62 & 59.31\\
		\hline
		ADDA & 59.03 & 55.48 & 53.71& 56.07 \\
		\hline
		Stylization & 67.42 & 60.45 & 58.62 & 62.16 \\
		\hline
		Contrastive learning & 66.18 & 58.60 & 56.38 & 60.39\\
		\hline
		Contrastive learning (enlarged unlabeled dataset) & 69.78 & 60.94 & 58.76 & 63.16 \\
		\hline
	\end{tabular}
\end{center}
\caption{Results of contrastive learning}
\end{table}

We can see from the table our method surpasses the baseline on most target dataset but the margin is not big enough. So we suspect that the unlabeled dataset is not big enough. The original unlabeled target dataset has only about 2-3k images, much smaller than the source dataset 20-40 k images.

\textbf{Increase the size of the unlabeled dataset} We increase the size of unlabeled dataset for two reasons: First, in most cases, unlabeled dataset are easy to build because no labels are needed and we don’t need experts to label them. Second, most un-supervised learning methods consume more samples compared to supervised learning methods. By increasing the size of the unlabeled dataset, we have more room for these algorithms. Our result is shown in table 4.2 after the size of unlabeled dataset is increased. From the table we can see that our method shows a big improvement on accuracies compared to our baselines, which means our method is very effective for this setting.

\subsection{Discussion}
This work is still ongoing and there are lots of things to be done. For example, the current clustering is still very slow, especially when the unlabeled dataset is very big. We need to find a faster clustering algorithm or reduce the number of clustering. Also we have another idea: combining domain-translation based methods and pseudo-based methods, which will take advantage of both methods and has a great potential to fully use the data and surpass all the methods.
\chapter{Future directions and open challenges} 
\label{Chapter5} 
In the next future years of my Phd research, there are several directions I want to explore:

\textbf{Cross-domain few shot learning with unlabeled data} It is still worth exploring. This is a more realistic setting and we should carefully study this problem and develop more algorithms for it. New research ideas can be drawn from UDA ReID or self-supervised learning.

\textbf{Compositional learning for few-shot learning} Current few-shot learning methods still heavily rely on learned features from the base dataset which does not and will not solve the few-shot learning. In my opinion, if we can find primitive features and then compose features from them. The data we need will be very little. The tasks can also be composed of primitive tasks if we can find the relations between different tasks.

\textbf{Causal inference for few shot learning} Causal inference is always an interesting topic which tries to find causal factors behind data. But it’s so different from machine learning and it has not been well studied by computer vision research yet. If we can fully utilize causal inference models, then we will be closer to artificial general intelligence and the data we need to train a model will also be much less.

\textbf{Analysis of common practice in few shot learning} Lots of common practice are only based on intuitions, which means they may be wrong practice. To move the research in this field forward, we need to carefully verify them by conducting thorough experiments or deriving theorems. For instance, why should we adopt episodic training? Why does pretraining help? 

\textbf{Continual few shot learning and open set learning} Continual learning means the model needs to learn a series of tasks continuously and the model can't forget previous tasks. This is an important feature of artificial general intelligence. Continual few shot learning is a excent testbed as few-shot learning is only conducted on small dataset. Open set learning means the model needs to deal with unseen classes during testing. This is related to meta learning and more realistic compared to the classic supervised learning.

To sum up, we need to address several challenges before we can achieve excellent performance on few-shot learning without the auxiliary dataset and become closer to artificial general intelligence.


\appendix 




\printbibliography[heading=bibintoc]

@incollection{NIPS2012_4824,
	title = {ImageNet Classification with Deep Convolutional Neural Networks},
	author = {Alex Krizhevsky and Sutskever, Ilya and Hinton, Geoffrey E},
	booktitle = {Advances in Neural Information Processing Systems 25},
	editor = {F. Pereira and C. J. C. Burges and L. Bottou and K. Q. Weinberger},
	pages = {1097--1105},
	year = {2012},
	publisher = {Curran Associates, Inc.},
	url = {http://papers.nips.cc/paper/4824-imagenet-classification-with-deep-convolutional-neural-networks.pdf}
}

@inproceedings{imagenet_cvpr09,
	AUTHOR = {Deng, J. and Dong, W. and Socher, R. and Li, L.-J. and Li, K. and Fei-Fei, L.},
	TITLE = {{ImageNet: A Large-Scale Hierarchical Image Database}},
	BOOKTITLE = {CVPR09},
	YEAR = {2009},
	BIBSOURCE = "http://www.image-net.org/papers/imagenet_cvpr09.bib"}

@article{devlin2018bert,
	title={Bert: Pre-training of deep bidirectional transformers for language understanding},
	author={Devlin, Jacob and Chang, Ming-Wei and Lee, Kenton and Toutanova, Kristina},
	journal={arXiv preprint arXiv:1810.04805},
	year={2018}
}

@article{brown2020language,
	title={Language models are few-shot learners},
	author={Brown, Tom B and Mann, Benjamin and Ryder, Nick and Subbiah, Melanie and Kaplan, Jared and Dhariwal, Prafulla and Neelakantan, Arvind and Shyam, Pranav and Sastry, Girish and Askell, Amanda and others},
	journal={arXiv preprint arXiv:2005.14165},
	year={2020}
}

@inproceedings{wang2018non,
	title={Non-local neural networks},
	author={Wang, Xiaolong and Girshick, Ross and Gupta, Abhinav and He, Kaiming},
	booktitle={Proceedings of the IEEE conference on computer vision and pattern recognition},
	pages={7794--7803},
	year={2018}
}

@article{he2015eep,
	title={Deep Residual Learning for Image Recognition},
	author={Kaiming He and Xiangyu Zhang and Shaoqing Ren and Jian Sun},
	year={2015},
	journal={arXiv preprint arXiv:1512.03385}
}

@article{chen2019,
	title={A Closer Look at Few-shot Classification},
	author={Wei-Yu Chen and Yen-Cheng Liu and Zsolt Kira and Yu-Chiang Frank Wang and Jia-Bin Huang},
	year={2019},
	journal={arXiv preprint arXiv:1904.04232}
}

@article{wang2019impleshot,
	title={SimpleShot: Revisiting Nearest-Neighbor Classification for Few-Shot
	Learning},
	author={Yan Wang and Wei-Lun Chao and Kilian Q. Weinberger and Laurens van der Maaten},
	year={2019},
	journal={arXiv preprint arXiv:1911.04623}
}

@article{tian2020ethinking,
	title={Rethinking Few-Shot Image Classification: a Good Embedding Is All You
	Need?},
	author={Yonglong Tian and Yue Wang and Dilip Krishnan and Joshua B. Tenenbaum and Phillip Isola},
	year={2020},
	journal={arXiv preprint arXiv:2003.11539}
}

@article{dhillon2019,
	title={A Baseline for Few-Shot Image Classification},
	author={Guneet S. Dhillon and Pratik Chaudhari and Avinash Ravichandran and Stefano Soatto},
	year={2019},
	journal={arXiv preprint arXiv:1909.02729}
}

@article{hospedales2020etalearning,
	title={Meta-Learning in Neural Networks: A Survey},
	author={Timothy Hospedales and Antreas Antoniou and Paul Micaelli and Amos Storkey},
	year={2020},
	journal={arXiv preprint arXiv:2004.05439}
}

@article{finn2017odelagnostic,
title={Model-Agnostic Meta-Learning for Fast Adaptation of Deep Networks},
author={Chelsea Finn and Pieter Abbeel and Sergey Levine},
year={2017},
journal={arXiv preprint arXiv:1703.03400}
}

@article{wang2019eneralizing,
	title={Generalizing from a Few Examples: A Survey on Few-Shot Learning},
	author={Yaqing Wang and Quanming Yao and James Kwok and Lionel M. Ni},
	year={2019},
	journal={arXiv preprint arXiv:1904.05046}
}

@article{rajeswaran2019etalearning,
	title={Meta-Learning with Implicit Gradients},
	author={Aravind Rajeswaran and Chelsea Finn and Sham Kakade and Sergey Levine},
	year={2019},
	journal={arXiv preprint arXiv:1909.04630}
}

@article{nichol2018n,
	title={On First-Order Meta-Learning Algorithms},
	author={Alex Nichol and Joshua Achiam and John Schulman},
	year={2018},
	journal={arXiv preprint arXiv:1803.02999}
}

@article{ravi2016optimization,
	title={Optimization as a model for few-shot learning},
	author={Ravi, Sachin and Larochelle, Hugo},
	year={2016}
}

@article{lee2019etalearning,
	title={Meta-Learning with Differentiable Convex Optimization},
	author={Kwonjoon Lee and Subhransu Maji and Avinash Ravichandran and Stefano Soatto},
	year={2019},
	journal={arXiv preprint arXiv:1904.03758}
}

@inproceedings{koch2015siamese,
	title={Siamese neural networks for one-shot image recognition},
	author={Koch, Gregory},
	year={2015}
}

@article{vinyals2016atching,
	title={Matching Networks for One Shot Learning},
	author={Oriol Vinyals and Charles Blundell and Timothy Lillicrap and Koray Kavukcuoglu and Daan Wierstra},
	year={2016},
	journal={arXiv preprint arXiv:1606.04080}
}

@article{sung2017earning,
	title={Learning to Compare: Relation Network for Few-Shot Learning},
	author={Flood Sung and Yongxin Yang and Li Zhang and Tao Xiang and Philip H. S. Torr and Timothy M. Hospedales},
	year={2017},
	journal={arXiv preprint arXiv:1711.06025}
}

@article{snell2017rototypical,
	title={Prototypical Networks for Few-shot Learning},
	author={Jake Snell and Kevin Swersky and Richard S. Zemel},
	year={2017},
	journal={arXiv preprint arXiv:1703.05175}
}

@inproceedings{santoro2016meta,
	title={Meta-learning with memory-augmented neural networks},
	author={Santoro, Adam and Bartunov, Sergey and Botvinick, Matthew and Wierstra, Daan and Lillicrap, Timothy},
	booktitle={International conference on machine learning},
	pages={1842--1850},
	year={2016}
}

@article{mishra2017,
	title={A Simple Neural Attentive Meta-Learner},
	author={Nikhil Mishra and Mostafa Rohaninejad and Xi Chen and Pieter Abbeel},
	year={2017},
	journal={arXiv preprint arXiv:1707.03141}
}

@article{munkhdalai2017eta,
	title={Meta Networks},
	author={Tsendsuren Munkhdalai and Hong Yu},
	year={2017},
	journal={arXiv preprint arXiv:1703.00837}
}

@article{gidaris2018ynamic,
	title={Dynamic Few-Shot Visual Learning without Forgetting},
	author={Spyros Gidaris and Nikos Komodakis},
	year={2018},
	journal={arXiv preprint arXiv:1804.09458}
}

@article{hariharan2016owshot,
	title={Low-shot Visual Recognition by Shrinking and Hallucinating Features},
	author={Bharath Hariharan and Ross Girshick},
	year={2016},
	journal={arXiv preprint arXiv:1606.02819}
}

@article{antoniou2017ata,
	title={Data Augmentation Generative Adversarial Networks},
	author={Antreas Antoniou and Amos Storkey and Harrison Edwards},
	year={2017},
	journal={arXiv preprint arXiv:1711.04340}
}

@article{wang2018owshot,
	title={Low-Shot Learning from Imaginary Data},
	author={Yu-Xiong Wang and Ross Girshick and Martial Hebert and Bharath Hariharan},
	year={2018},
	journal={arXiv preprint arXiv:1801.05401}
}

@article{zhang2019ewshot,
	title={Few-Shot Learning via Saliency-guided Hallucination of Samples},
	author={Hongguang Zhang and Jing Zhang and Piotr Koniusz},
	year={2019},
	journal={arXiv preprint arXiv:1904.03472}
}

@article{chen2019mage,
	title={Image Deformation Meta-Networks for One-Shot Learning},
	author={Zitian Chen and Yanwei Fu and Yu-Xiong Wang and Lin Ma and Wei Liu and Martial Hebert},
	year={2019},
	journal={arXiv preprint arXiv:1905.11641}
}

@article{cubuk2018utoaugment,
	title={AutoAugment: Learning Augmentation Policies from Data},
	author={Ekin D. Cubuk and Barret Zoph and Dandelion Mane and Vijay Vasudevan and Quoc V. Le},
	year={2018},
	journal={arXiv preprint arXiv:1805.09501}
}

@article{zhang2017ixup,
	title={mixup: Beyond Empirical Risk Minimization},
	author={Hongyi Zhang and Moustapha Cisse and Yann N. Dauphin and David Lopez-Paz},
	year={2017},
	journal={arXiv preprint arXiv:1710.09412}
}

@article{verma2018anifold,
	title={Manifold Mixup: Better Representations by Interpolating Hidden States},
	author={Vikas Verma and Alex Lamb and Christopher Beckham and Amir Najafi and Ioannis Mitliagkas and Aaron Courville and David Lopez-Paz and Yoshua Bengio},
	year={2018},
	journal={arXiv preprint arXiv:1806.05236}
}

@article{huang2017rbitrary,
	title={Arbitrary Style Transfer in Real-time with Adaptive Instance
	Normalization},
	author={Xun Huang and Serge Belongie},
	year={2017},
	journal={arXiv preprint arXiv:1703.06868}
}

@misc{DataAugm34:online,
	author = {},
	title = {Data Augmentation: How to use Deep Learning when you have Limited Data},
	howpublished = {\url{https://www.kdnuggets.com/2018/05/data-augmentation-deep-learning-limited-data.html}},
	month = {},
	year = {},
	note = {(Accessed on 10/23/2020)}
}

@misc{Howtosty65:online,
	author = {},
	title = {How to style transfer your own images - GoDataDriven},
	howpublished = {\url{https://godatadriven.com/blog/how-to-style-transfer-your-own-images/}},
	month = {},
	year = {},
	note = {(Accessed on 10/23/2020)}
}

@article{vaswani2017ttention,
	title={Attention Is All You Need},
	author={Ashish Vaswani and Noam Shazeer and Niki Parmar and Jakob Uszkoreit and Llion Jones and Aidan N. Gomez and Lukasz Kaiser and Illia Polosukhin},
	year={2017},
	journal={arXiv preprint arXiv:1706.03762}
}

@article{dosovitskiy2020n,
	title={An Image is Worth 16x16 Words: Transformers for Image Recognition at
	Scale},
	author={Alexey Dosovitskiy and Lucas Beyer and Alexander Kolesnikov and Dirk Weissenborn and Xiaohua Zhai and Thomas Unterthiner and Mostafa Dehghani and Matthias Minderer and Georg Heigold and Sylvain Gelly and Jakob Uszkoreit and Neil Houlsby},
	year={2020},
	journal={arXiv preprint arXiv:2010.11929}
}

@inproceedings{chen2020generative,
	title={Generative Pretraining from Pixels},
	author={Chen, Mark and Radford, Alec and Child, Rewon and Wu, Jeff and Jun, Heewoo and Dhariwal, Prafulla and Luan, David and Sutskever, Ilya},
	booktitle={Proceedings of the 37th International Conference on Machine Learning},
	year={2020}
}

@misc{Attentio84:online,
	author = {},
	title = {Attention? Attention!},
	howpublished = {\url{https://lilianweng.github.io/lil-log/2018/06/24/attention-attention.html}},
	month = {},
	year = {},
	note = {(Accessed on 10/24/2020)}
}

@article{zhang2020eepemd,
	title={DeepEMD: Differentiable Earth Mover's Distance for Few-Shot Learning},
	author={Chi Zhang and Yujun Cai and Guosheng Lin and Chunhua Shen},
	year={2020},
	journal={arXiv preprint arXiv:2003.06777}
}

@article{raghu2019apid,
	title={Rapid Learning or Feature Reuse? Towards Understanding the Effectiveness
	of MAML},
	author={Aniruddh Raghu and Maithra Raghu and Samy Bengio and Oriol Vinyals},
	year={2019},
	journal={arXiv preprint arXiv:1909.09157}
}

@article{tseng2020rossdomain,
	title={Cross-Domain Few-Shot Classification via Learned Feature-Wise
	Transformation},
	author={Hung-Yu Tseng and Hsin-Ying Lee and Jia-Bin Huang and Ming-Hsuan Yang},
	year={2020},
	journal={arXiv preprint arXiv:2001.08735}
}

@article{guo2019,
	title={A Broader Study of Cross-Domain Few-Shot Learning},
	author={Yunhui Guo and Noel C. Codella and Leonid Karlinsky and James V. Codella and John R. Smith and Kate Saenko and Tajana Rosing and Rogerio Feris},
	year={2019},
	journal={arXiv preprint arXiv:1912.07200}
}

@inproceedings{gao2020deep,
	title={Deep Clustering for Domain Adaptation},
	author={Gao, Boyan and Yang, Yongxin and Gouk, Henry and Hospedales, Timothy M},
	booktitle={ICASSP 2020-2020 IEEE International Conference on Acoustics, Speech and Signal Processing (ICASSP)},
	pages={4247--4251},
	year={2020},
	organization={IEEE}
}

@article{chang2018isjoint,
	title={Disjoint Label Space Transfer Learning with Common Factorised Space},
	author={Xiaobin Chang and Yongxin Yang and Tao Xiang and Timothy M. Hospedales},
	year={2018},
	journal={arXiv preprint arXiv:1812.02605}
}

@article{khodadadeh2018nsupervised,
	title={Unsupervised Meta-Learning For Few-Shot Image Classification},
	author={Siavash Khodadadeh and Ladislau Bölöni and Mubarak Shah},
	year={2018},
	journal={arXiv preprint arXiv:1811.11819}
}

@article{hsu2018nsupervised,
	title={Unsupervised Learning via Meta-Learning},
	author={Kyle Hsu and Sergey Levine and Chelsea Finn},
	year={2018},
	journal={arXiv preprint arXiv:1810.02334}
}

@article{zhu2017npaired,
	title={Unpaired Image-to-Image Translation using Cycle-Consistent Adversarial
	Networks},
	author={Jun-Yan Zhu and Taesung Park and Phillip Isola and Alexei A. Efros},
	year={2017},
	journal={arXiv preprint arXiv:1703.10593}
}

@article{geirhos2018magenettrained,
	title={ImageNet-trained CNNs are biased towards texture; increasing shape bias
	improves accuracy and robustness},
	author={Robert Geirhos and Patricia Rubisch and Claudio Michaelis and Matthias Bethge and Felix A. Wichmann and Wieland Brendel},
	year={2018},
	journal={arXiv preprint arXiv:1811.12231}
}

@article{ge2020elfpaced,
	title={Self-paced Contrastive Learning with Hybrid Memory for Domain Adaptive
	Object Re-ID},
	author={Yixiao Ge and Feng Zhu and Dapeng Chen and Rui Zhao and Hongsheng Li},
	year={2020},
	journal={arXiv preprint arXiv:2006.02713}
}

@article{peng2018oment,
	title={Moment Matching for Multi-Source Domain Adaptation},
	author={Xingchao Peng and Qinxun Bai and Xide Xia and Zijun Huang and Kate Saenko and Bo Wang},
	year={2018},
	journal={arXiv preprint arXiv:1812.01754}
}

@article{tzeng2017dversarial,
	title={Adversarial Discriminative Domain Adaptation},
	author={Eric Tzeng and Judy Hoffman and Kate Saenko and Trevor Darrell},
	year={2017},
	journal={arXiv preprint arXiv:1702.05464}
}

@inproceedings{ester1996density,
	title={A density-based algorithm for discovering clusters in large spatial databases with noise.},
	author={Ester, Martin and Kriegel, Hans-Peter and Sander, J{\"o}rg and Xu, Xiaowei and others},
	booktitle={Kdd},
	volume={96},
	number={34},
	pages={226--231},
	year={1996}
}

@article{chen2020,
	title={A Simple Framework for Contrastive Learning of Visual Representations},
	author={Ting Chen and Simon Kornblith and Mohammad Norouzi and Geoffrey Hinton},
	year={2020},
	journal={arXiv preprint arXiv:2002.05709}
}

@article{qi2017owshot,
    title={Low-Shot Learning with Imprinted Weights},
    author={Hang Qi and Matthew Brown and David G. Lowe},
    year={2017},
    journal={arXiv preprint arXiv:1712.07136}
}

@article{gatys2015,
    title={A Neural Algorithm of Artistic Style},
    author={Leon A. Gatys and Alexander S. Ecker and Matthias Bethge},
    year={2015},
    journal={arXiv preprint arXiv:1508.06576}
}

@article{dumoulin2016,
    title={A Learned Representation For Artistic Style},
    author={Vincent Dumoulin and Jonathon Shlens and Manjunath Kudlur},
    year={2016},
    journal={arXiv preprint arXiv:1610.07629}
}

\end{document}